\documentclass[10pt,letterpaper]{article}

\usepackage{ccn}
\usepackage{pslatex}
\usepackage{apacite}
\usepackage{graphicx}
\usepackage{caption}
\usepackage{natbib}
\usepackage{float}

\usepackage{hyperref}       
\usepackage{url}            
\usepackage{booktabs}       
\usepackage{amsfonts}       
\usepackage{nicefrac}       
\usepackage{microtype}      
\usepackage{xcolor}         
\usepackage{amsmath}
\usepackage{amssymb}
\usepackage{mathtools}
\usepackage{amsthm}
\usepackage{graphicx}
\usepackage{svg}
\usepackage[labelfont=bf, margin=0.8cm, font=sf]{caption}

\title{Revisiting the Role of Relearning in Semantic Dementia}
 
\author{{\large \bf Devon Jarvis (devon.jarvis@wits.ac.za)} \\ University of the Witwatersrand, 1 Jan Smuts Ave,\\ Braamfontein, Johannesburg, 2000, South Africa \\ and \\ Gatsby Computational Neuroscience Unit \& Sainsbury Wellcome Centre, University College London, 25 Howland Street,\\ London W1T 4JG, United Kingdom
  \AND {\large \bf Verena Klar (verena.klar@psy.ox.ac.uk)} \\ Department of Experimental Psychology, Anna Watts Building, Woodstock Rd,\\ Oxford OX2 6GG United Kingdom
  \AND {\large \bf Richard Klein (richard.klein@wits.ac.za)} \\ University of the Witwatersrand, 1 Jan Smuts Ave,\\ Braamfontein, Johannesburg, 2000, South Africa
  \AND {\large \bf Benjamin Rosman (Benjamin.Rosman1@wits.ac.za)} \\ University of the Witwatersrand, 1 Jan Smuts Ave,\\ Braamfontein, Johannesburg, 2000, South Africa
  \AND {\large \bf Andrew Saxe (a.saxe@ucl.ac.uk)} \\ Gatsby Computational Neuroscience Unit \& Sainsbury Wellcome Centre, University College London, 25 Howland Street,\\ London W1T 4JG, United Kingdom}

\begin{document}

\maketitle
\phantom{.}
\newpage
\phantom{.}
\newpage
\section{Abstract}
{
\bf
\looseness=-1
Patients with semantic dementia (SD) present with remarkably consistent atrophy of neurons in the anterior temporal lobe and behavioural impairments, such as graded loss of category knowledge. While relearning of lost knowledge has been shown in acute brain injuries such as stroke, it has not been widely supported in chronic cognitive diseases such as SD. Previous research has shown that deep linear artificial neural networks exhibit stages of semantic learning akin to humans. Here, we use a deep linear network to test the hypothesis that relearning during disease progression rather than particular atrophy cause the specific behavioural patterns associated with SD. After training the network to generate the common semantic features of various hierarchically organised objects, neurons are successively deleted to mimic atrophy while retraining the model. The model with relearning and deleted neurons reproduced errors specific to SD, including prototyping errors and cross-category confusions. This suggests that relearning is necessary for artificial neural networks to reproduce the behavioural patterns associated with SD in the absence of \textit{output} non-linearities. Our results support a theory of SD progression that results from continuous relearning of lost information. Future research should revisit the role of relearning as a contributing factor to cognitive diseases.
}
\begin{quote}
\small
\textbf{Keywords:} 
Semantic Dementia, Linear Neural Networks, Relearning, Anterior Temporal Lobes, Representation Learning
\end{quote}
\section{Introduction}

\begin{figure}[t]
  \captionsetup{margin=0.0in, font=sf}
  \begin{minipage}{\linewidth}
  \centering
    \includegraphics[width=1.0\linewidth]{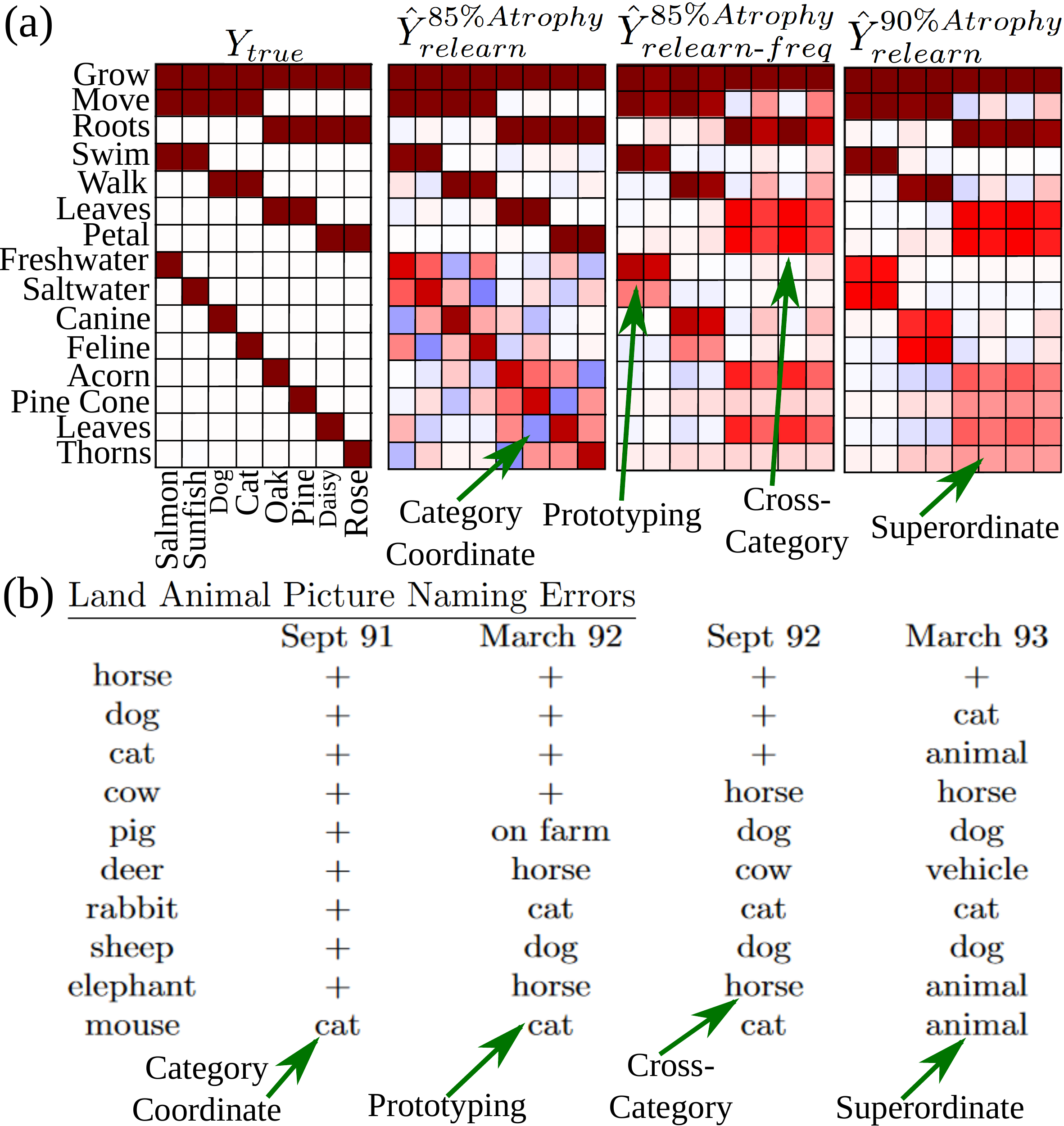}
  \end{minipage}
  \caption{\looseness=-1 As with a patient with SD ((b) taken from \citet{hodges1995charting}), the model (a) makes category coordinate errors first and progresses to cross-category and superordinate errors. Additionally, if odd data points are shown twice as often during relearning ($\hat{Y}_{relearn\text{-}freq}$) then the more frequently seen features will dominate the representations. As a result, the less frequent objects are mistaken for the prototypical ones when making category coordinate and cross-category errors. Thus, a linear network with relearning reproduces the pattern of errors and prototyping effect associated with SD in humans.}
  \label{fig:errors}
\end{figure}

\looseness=-1
Patients with semantic dementia (SD) show a particular progression of behavioural impairments which result from the disease \citep{hodges1995charting} (examples shown in Figure \ref{fig:errors}(b)). Specifically, in SD errors of hierarchically organised semantic knowledge occurs first for fine grained distinctions at the bottom of a hierarchy (termed category coordinate errors) and progresses upwards until errors are made for more semantically distinct objects (cross-category errors) and fine grained distinctions are forgotten (superordinate errors) \citep{jefferies2006semantic}. There is also a distinct typicality effect where patients are better able to identify frequently seen objects of a category with very typical features. This results in prototyping, where all objects of a category are named as the most stereotypical object in the category. SD has been linked to bilateral atrophy of the anterior temporal lobe (ATL). This finding in conjunction with converging evidence from brain imaging \citep{devlin2000susceptibility,visser2010semantic} has led to the region being regarded as an amodal semantic hub \citep{jackson2021reverse}. 

\looseness=-1
In this work, we explore the ability of simple linear neural network models to reproduce the pattern of behavioural impairment from SD. Such a model has been used successfully in the past to reproduce regularities in the development of semantic cognition during early childhood \citep{saxe2019mathematical}. The networks hidden layer can be thought to represent the ATL. We test the hypothesis that disruption (atrophy) to the hidden layer itself is not sufficient as a model of SD, but that progression of impairment is better explained by a combination of atrophy and relearning.

\setlength{\belowcaptionskip}{-15pt}
\setlength{\belowcaptionskip}{0pt}

\section{Methods}


\setlength{\belowcaptionskip}{-7pt}
\begin{figure*}[ht!]
    \centering
    \includegraphics[width=0.92\linewidth]{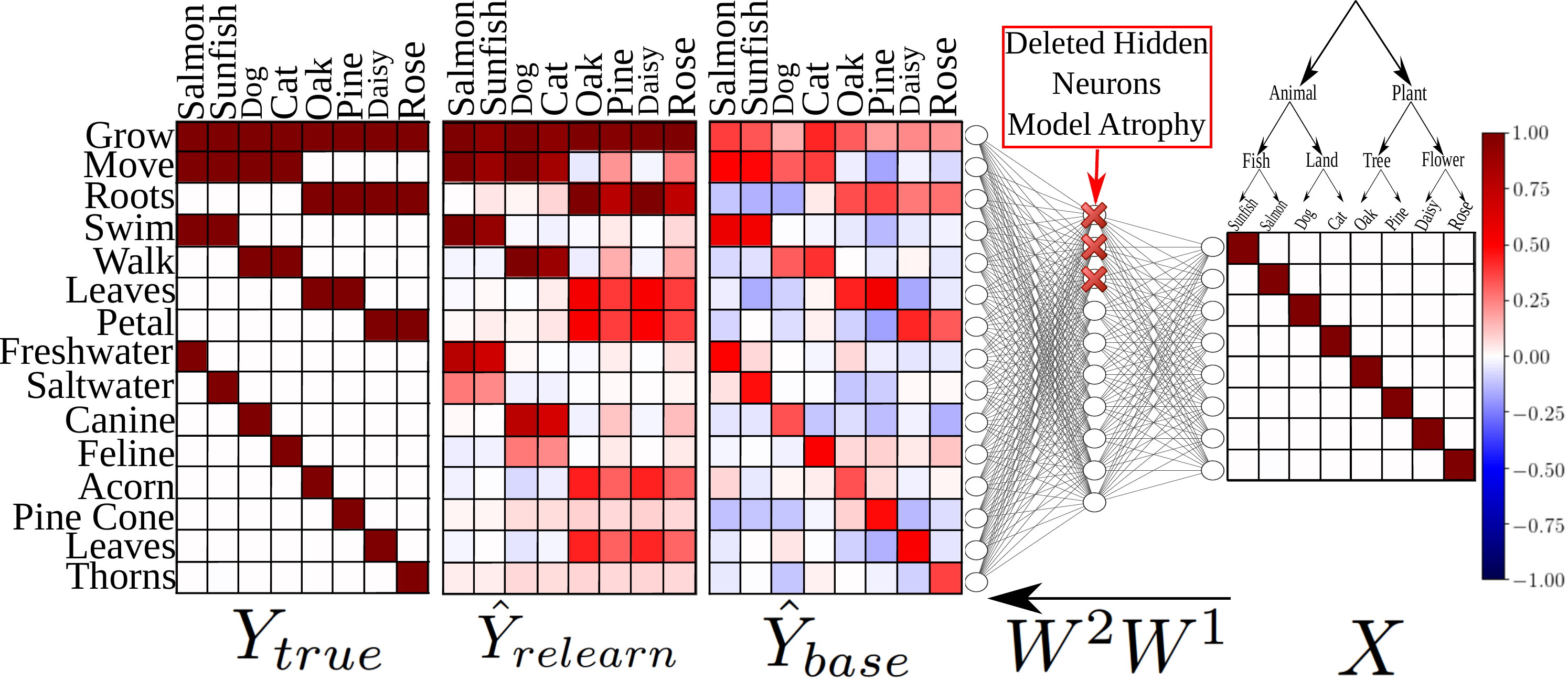}
    \captionsetup{font=sf}
    \caption{\looseness=-1  Setup and Primary Results: The linear network ($W^2W^1$) learns to map hierarchically structured data ($X$) to the corresponding features ($Y_{true}$). We model SD by deleting hidden neurons after training is complete. We compare two models: one with no relearning ($\hat{Y}_{base}$) and one with relearning ($\hat{Y}_{relearn}$). A model with no relearning loses information across all levels of the hierarchy at once, contradicting the patterns associated to SD. A model with relearning loses specific feature information before general features, consistent with what is expected of SD.}
    \label{fig:relearn}
\end{figure*}
\setlength{\belowcaptionskip}{0pt}

\subsubsection{Model}
\looseness=-1
We trained a linear neural network to describe the features of an object in a semantically hierarchical dataset, where features are shared based on how close objects are in the hierarchy. In contrast to prior approaches \citep{rogers2004semantic}, we do not rely on \textbf{output} non-linearities (specifically thresholding) to reproduce SD with our model. Instead, our network outputs real values representing the network's confidence in the presence of each feature. This allows for more direct comparison with human behavioural tasks with fine-grained actions or responses, such as picture selection tasks \citep{jefferies2006semantic} or object drawing tasks \citep{bozeat2003duck}.

\setlength{\belowcaptionskip}{-7pt}
\begin{table}[h!]
\centering
\begin{tabular}{|c|c|c|} 
 \hline
 Model & \multicolumn{1}{|p{2cm}|}{\centering Level 1 Error\\ (Percentage)} & \multicolumn{1}{|p{2cm}|}{\centering Level 4 Error\\ (Percentage)} \\
 \hline\hline
 Linear; $n=0$ & 58.25 & 79.88 \\ 
 \hline
 Linear; $n=200$ & 0.0 & 62.50\\
 \hline
 ReLU; $n=0$ & 51.50 & 84.50 \\ 
 \hline
 ReLU; $n=200$ & 2.75 & 62.50 \\
 \hline
\end{tabular}
\captionsetup{font=sf, width=0.95\linewidth, skip=7pt}
\caption{\looseness=-1 Relearning is necessary for models with linear and ReLU activations on the hidden layer to lose fine grained features before higher-level features due to atrophy in the hidden layer. Note that the output layer is still linear for all models (error is as a percentage compared to a naive model to aid comparison between hierarchy layers).}
\label{tab:nonlin}
\end{table}
\setlength{\belowcaptionskip}{0pt}

\looseness=-1
\subsubsection{Training regime}
We first trained the network to convergence using (full-batch) gradient descent from small initial weights. This results in the network learning weights which identify the hierarchical nature of the input and map each distinction in the hierarchy to a corresponding set of output features. After convergence we deleted neurons in the hidden layer to model atrophy and retrained the model for $n$ epochs after each deletion. This was repeated until no hidden neurons remained. A summary of the hierarchical dataset, network and findings are shown in Figure \ref{fig:relearn}.

\section{Results}
\looseness=-1
The behavioural impacts of atrophy depends on how information is distributed in the hidden neurons and are represented by the errors made by the network. Our results show that a linear network is unable to reproduce the behaviour patterns of SD without relearning ($n = 0$) as the loss of semantic knowledge occurs for all levels of the hierarchy at once (see $\hat{Y}_{base}$ in Figure~\ref{fig:errors}(a) and the first row of Table \ref{tab:nonlin}). However, if relearning is used ($n > 0$) the network reproduces these patterns (see $\hat{Y}_{relearn}$ in Figure~\ref{fig:errors}(a) and the second row of Table \ref{tab:nonlin}). Table \ref{tab:nonlin} also shows that the same effect can be seen when non-linearity is used on the hidden layer of the network. 

\looseness=-1
Specifically, the network makes errors on the fine grained semantic features after fewer neurons are deleted (less atrophy) than for higher-level semantic features. Thus, category coordinate errors occur first. As the atrophy increases the network makes errors on semantic features for higher levels of the hierarchy (cross-category errors) and produces near $0$ output for fine grained features which are forgotten (superordinate errors). An example of the pattern of errors from the model is shown in Figure~\ref{fig:errors}(a) and can be compared to the example progression of the human patient from \citet{hodges1995charting}. Finally, Figure~\ref{fig:errors}(a) also demonstrates that our model is sensitive to the frequency that it encounters features with. The model more rapidly forgets the semantic features which are encountered less often and makes errors by providing the features for the most stereotypical object at a given level of the hierarchy. Thus, our model also reproduces the prototyping effect observed in human SD patients \citep{bozeat2003duck}.

\section{Discussion}
We have demonstrated that the typical progression of behaviour impairment from SD: category coordinate errors leading to cross category and superordinate errors; only emerges with relearning in an artificial neural network with linear output producing semantic features for hierarchically structured objects. Our results suggest that it is relearning after atrophy -- the brain's attempt to adapt to the disease -- rather than atrophy itself which causes patterns of memory degradation in SD. A similar role of relearning has been established in herpes simplex virus encephalitis \citep{ralph2007neural}, but remains unexamined in other chronic cognitive diseases. Our results call for consideration of the role of relearning as a primary cause of the known behavioural patterns in these diseases.

\section{Acknowledgements}
This work was supported by the Commonwealth Scholarship and Google PhD Fellowship to DJ, the ESRC ES/P000649/1 and New College 1379 Old Members Scholarship to VK, a Sir Henry Dale Fellowship from the Wellcome Trust and Royal Society (216386/Z/19/Z) to A.S., and the Sainsbury Wellcome Centre Core Grant from Wellcome (219627/Z/19/Z) and the Gatsby Charitable Foundation (GAT3755). A.S. and B.R. are CIFAR Azrieli Global Scholars in the Learning in Machines \& Brains program.

\bibliographystyle{apacite}

\setlength{\bibleftmargin}{.125in}
\setlength{\bibindent}{-\bibleftmargin}

\bibliography{ccn_style}

\end{document}